\documentclass[11pt]{article}
\PassOptionsToPackage{table}{xcolor}

\usepackage[final]{acl}

\usepackage{soul}

\usepackage{times}
\usepackage{latexsym}

\usepackage[T1]{fontenc}

\usepackage[utf8]{inputenc}

\usepackage{microtype}

\usepackage{inconsolata}

\usepackage{graphicx}
\usepackage{multirow}
\usepackage{booktabs}
\usepackage{makecell}
\usepackage{algorithm}
\usepackage{algorithmic}
\usepackage{amsmath}
\usepackage{tabularx}
\usepackage{enumitem}
\usepackage{breqn}
\usepackage{arydshln}
\newcolumntype{C}{>{\raggedright\arraybackslash}X}
\title{Auto Review: Second Stage Error Detection for Highly Accurate Information Extraction from Phone Conversations}

\author{Ayesha Qamar,
  Arushi Raghuvanshi, Conal Sathi, \and Youngseo Son  \\
  Infinitus Systems, Inc. \\
  \{ayesha, arushi, conal, youngseo.son\}@infinitus.ai
  }

\begin{document}
\maketitle
\begin{abstract}

\end{abstract}
Automating benefit verification phone calls saves time in healthcare and helps patients receive treatment faster. It is critical to obtain highly accurate information in these phone calls, as it can affect a patient's healthcare journey.
Given the noise in phone call transcripts, we have a two-stage system that involves a post-call review phase for potentially noisy fields, where human reviewers manually verify the extracted data—a labor-intensive task. To automate this stage, we introduce \textbf{\textit{Auto Review}}, which significantly reduces manual effort while maintaining a high bar for accuracy.
This system, being highly reliant on call transcripts, suffers a performance bottleneck due to automatic speech recognition (ASR) issues.
This problem is further exacerbated by the use of domain-specific jargon in the calls.
In this work, we propose a second-stage postprocessing pipeline for accurate information extraction. 
We improve accuracy by using multiple ASR alternatives and a pseudo-labeling approach that does not require manually corrected transcripts. Experiments with general-purpose large language models and feature-based model pipelines demonstrate substantial improvements in the quality of corrected call transcripts, thereby enhancing the efficiency of \textbf{\textit{Auto Review}}. 

\section{Introduction}
\begin{figure}[t] 
    \centering
    \captionsetup{font=footnotesize}
    \includegraphics[width=0.5\textwidth]{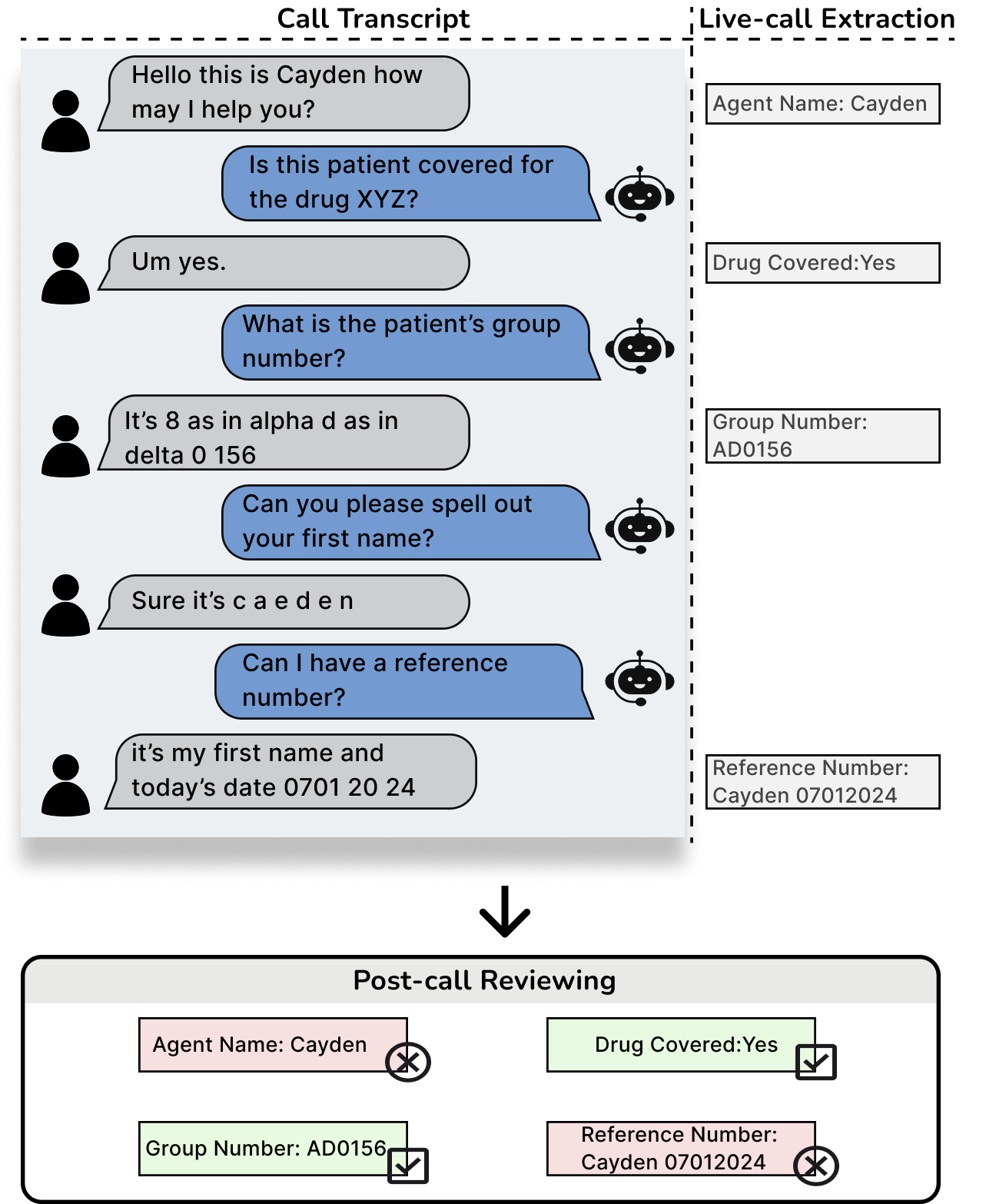}
    \captionsetup{font=footnotesize}
    \caption{An excerpt from a dummy chat, along with the field values extracted during the call, is passed to the post-call reviewing module for verification. The noisy ASR transcripts can contribute to errors in the extracted data; this is exacerbated for domain-specific jargon such as group number and rare agent names.}
    \label{fig:example_snapshot}
\end{figure}

A key use case for Conversational AI systems in industry is collecting information~\citep{gnewuch2017towards}. One critical application is healthcare benefit verification, where information about a patient’s insurance coverage is gathered from an insurance company over the phone.
These extracted values, such as patient group numbers and drug coverage details, are essential for treatment approval and directly impact a patient’s healthcare journey~\citep{buker2023financial}. Given the high-stakes nature of this task, ensuring the accuracy of extracted data is crucial.

While extensive research has focused on conversation navigation techniques—such as intent prediction, slot filling, and dialogue state tracking~\citep{mctear2022conversational}—there has been comparatively less emphasis on ensuring the accuracy of extracted information in AI-driven conversations with task-specific context. In real-world applications, automated phone call outputs often contain errors due to ASR challenges, including background noise, domain-specific jargon, and complex alphanumeric sequences.
To maintain data reliability, it is crucial to incorporate automated error correction methods or human-in-the-loop verification where necessary.
Unlike prior work that focuses on ASR error correction for grammatical mistakes, our goal is to improve the accuracy of extracted informational fields. 
Since creating datasets for ASR error correction is time-consuming and labor-intensive, we propose using a pseudo-labeling technique with Large Language Models (LLMs).

Given the real-time constraints of compute and latency during live calls, we introduce \textbf{Auto Review}, a two-stage pipeline that enhances post-call information extraction. The first stage involves a conversational AI system that navigates live calls and extracts key field values. However, it does not guarantee that the extracted values are highly accurate. The second stage performs an automated review, flagging potential errors for human review or approving the accurate values. This second stage significantly reduces manual human review time while maintaining high accuracy.

We evaluate LLMs as a reviewing agent in two distinct settings: direct verification, where a model determines whether an extracted field value in the first stage is correct, and direct extraction, where a model identifies the correct value directly from the transcript. We compare multiple LLMs and feature-based models, analyzing their trade-offs in precision, recall, and computational efficiency.

The main contributions of this paper can be summarized as:
\begin{itemize}
    \setlength{\itemsep}{0.01em}
    \item We introduce a two-stage pipeline for accurate and efficient information extraction in the healthcare benefit verification domain. This approach saves human review time while ensuring high accuracy in the final outputs delivered to clients.
    \item To address domain-specific errors in ASR transcripts, we propose a pseudo-label generation technique leveraging LLMs.
    \item We conduct a comprehensive evaluation of LLMs for information verification in both generative and discriminative settings, analyzing the trade-offs between the two approaches.
\end{itemize}

\section{Related Work}
\paragraph{ASR Error Correction} Most research on ASR error detection and correction focuses on grammatical mistakes~\citep{li2024detection, ma2023n}. ~\citet{loem2023exploring} demonstrated that GPT-3, in zero-shot and few-shot settings, can perform grammatical error correction. ~\citet{davis2024prompting} used LLM prompting techniques to address grammatical issues, while ~\citet{wang2024improving} combined rule-based methods with generative models to introduce artificial errors that mimic real-world patterns. ~\citet{shen2022mask} highlighted how the scarcity of errors in training data limits a model’s ability to correct them effectively.
Unlike these approaches, our focus is on correcting informational fields rather than grammatical issues. We leverage domain-specific context and frequent ASR error patterns to improve accuracy in benefit verification.

Previous work has focused on correcting named entity errors in ASR text. For instance, \citet{pusateri2024retrieval} use a retrieval-augmented approach, while \citet{discriminative-entity-aware-language} leverage external knowledge sources like knowledge graphs. However, in our healthcare phone conversations, sensitive and context-dependent information (e.g., personal health data) is often not available in public knowledge bases and can only be captured live during the call.

Many studies use supervised fine-tuning as a post-processing step to reduce ASR errors~\citep{errattahi2016automatic, radhakrishnan2023whispering}. Some approaches~\citep{ebadi2024extracting} avoid relying on manually corrected transcripts by using the inherent knowledge of LLMs to correct errors. In contrast, we don't have manually corrected transcripts, and few-shot LLMs were ineffective, as they haven't been exposed to our domain-specific data during pre-training.

\paragraph{Output Extraction} Dialogue state tracking (DST) in task-oriented dialogues involves intent recognition, which can be viewed as output extraction based on the user turns~\citep{li2024large}. This process fills predefined slot-value pairs according to the domain and task requirements. In healthcare benefit verification, this translates to extracting specific fields necessary to confirm patient benefits~\citep{feng2023towards}.
Retrieval-augmented strategies have been explored for DST~\citep{king2023diverse}, and LLMs have been applied to intent and entity extraction for live conversations~\citep{luo2024zero}. While our first-stage live call system incorporates elements of these approaches, it does not achieve the required accuracy given our healthcare-specific constraints on latency and compute resources. To address this, we introduce a second-stage system that refines outputs in a post-processing step, improving overall accuracy.

\section{Two Stage Pipeline for Highly Accurate Information Extraction}
Our automation pipeline for verifying patient insurance benefits involves two stages. First, a live-call conversational AI model engages with an insurance representative to collect the necessary benefit information. Second, an auto-review AI model validates the collected data based on the full call context, patient details, and domain knowledge.

The goal is to ensure the accuracy of the information and automate healthcare processes. In cases where the data may be uncertain, a human is brought in for review. For high-confidence fields, we can automatically approve the data, significantly reducing human involvement and improving operational efficiency without compromising quality.

\begin{figure}
    \captionsetup{font=footnotesize}
    \includegraphics[width=0.5\textwidth]{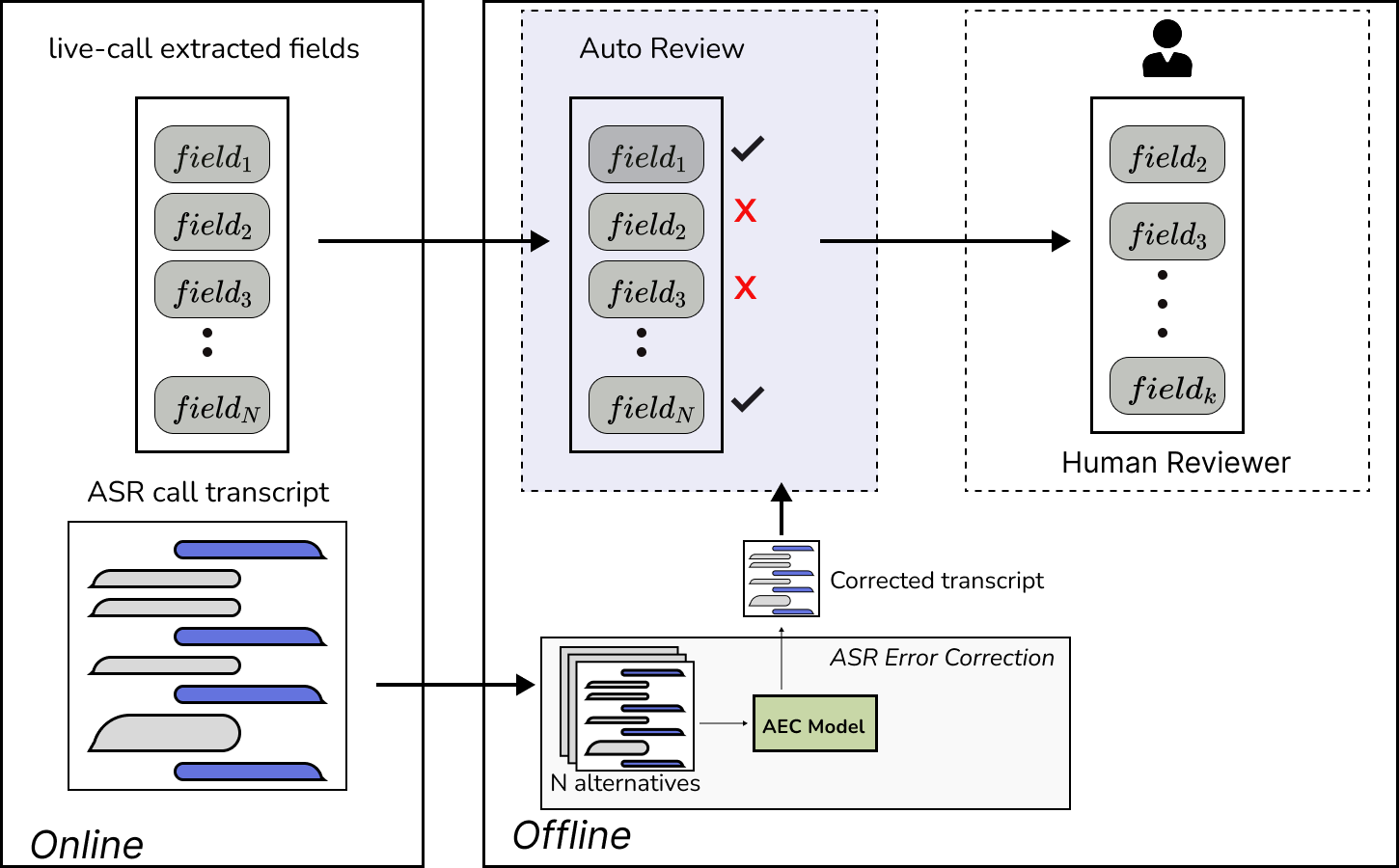}
    \captionsetup{font=footnotesize}
    \caption{The \textit{auto review} pipeline consists of an online and an offline component. The fields that do not get auto-approved are passed to a human reviewer for correction.}
    \label{fig:overview_components}
\end{figure}

In the second stage, the auto-review AI models verify the accuracy of the information collected. We define auto-reviewing as the process of assessing whether each extracted value from a call transcript is correct. As shown in the conversation snapshot in ~\autoref{fig:example_snapshot}, some information may be updated or corrected during the call.

To support large-scale industrial deployment, we prioritized cost-effective model design, considering trade-offs between model complexity and performance. Our objective is to deploy efficient and scalable models that maintain comparable performance to larger alternatives, as long as differences are not statistically significant.
The models evaluated in this paper represent a simplified component of a broader production pipeline used in our industrial setting.

\section{Data Description}
\begin{table}
\centering
\small
\begin{tabular}{l|l|ll}
    \hline
    \textbf{Field} & \textbf{Error Rates} & \textbf{Mean Edit} & \textbf{STDV} \\
    \hline
    Agent Name & 10.80\% & 3.23 & 2.89 \\
    Reference Number & 12.90\% & 7.05 & 6.43 \\
    Group Number & 9.80\% & 3.76 & 7.76 \\
    \hline
\end{tabular}
\captionsetup{font=footnotesize}
\caption{Error rates denote the ratio of incorrectly extracted live-call values for each field. Mean edit and STD denote mean and standard deviations of edit distances of live-call extracted values that contain errors.}
\label{tab:error_stats}
\end{table}

We collected 9,456 benefit verification calls between February and July 2024 for our experiments. Calls from February 1st to July 3rd were used for training, calls from July 5th for validation, and calls from July 10th to 12th for evaluation\footnote{No calls were collected over the weekend.}. The dataset details are given in~\autoref{tab:dataset}. The dataset includes call audio, ASR transcripts, extracted field values, and human-verified gold field values.

The field values in our healthcare domain include alphanumeric strings (e.g., insurance agent name, patient group number), booleans (e.g., medication coverage), and dates (e.g., effective dates of insurance plans). Alphanumeric fields typically exhibit the highest error rates due to ASR mistranscriptions caused by homophones, background noise, and similar-sounding names.
We focus on alphanumeric fields for three reasons: 1) they have the highest correction rates, 2) they vary greatly in value, and 3) they are most prone to ASR errors. Therefore, we discuss three key alphanumeric fields with the highest correction rates: Agent Name, Reference Number, and Group Number\footnote{Multimodal LLMs performed poorly when directly extracting from call audio recordings (see \ref{sec:prelim_audio_gemini}).}.
The first-stage conversational AI models were generally accurate, with target output fields having an error correction rate of 10-13\%, and their mean edit distances ranging from 3.23 to 7.05 (see \autoref{tab:error_stats}).

\begin{table}
\centering
\small
\begin{tabular}{l|l|lll}
\hline
\textbf{Dataset Type} & \textbf{Calls} & \textbf{AVG} & \textbf{STDV}\\
\hline
Train & 6,652  & 907  & 316.09 \\ 
Validation & 383 & 926 & 329.26 \\ 
Test & 2,260 & 939 & 356.79 \\ \hline 
\end{tabular}
\captionsetup{font=footnotesize}
\caption{Patients benefit verification phone calls. AVG: average number of words, STDV: standard deviation.}
\label{tab:dataset}
\end{table}

\section{Auto-Review Model}

We developed two primary approaches for automatically reviewing benefit information, both of which take the call transcript as input. The first, \textbf{\textit{Direct Extraction}}, extracts the field values, while the second, \textbf{\textit{Direct Verification}}, uses the live-call values and determines, in a discriminative setting, whether they are correct.

\subsection{Direct Verification}

In this approach, both the transcript and the live-call field value are provided as input. The \textit{live-call} value is defined as the field value extracted by our real-time system, which may also involve human in the loop. This setting is akin to binary classification.

\textit{Input: [Transcript][Live-call Extracted Field Value] Is the field value correct? Output: Yes/No}

\subsection{Direct Extraction} 
Here, the model receives the call transcript along with the field name and is tasked with extracting the relevant value from the transcript. The value extracted in this setting is referred to as the \textit{post-call} value.

\textit{Input: [Transcript] What is the field value? Output: Post-call Extracted Field Value}

After the extraction, we convert the task back to a review process by comparing the extracted field value with the live-call field value. If the live-call field value matches the post-call extracted value, we consider it to be correct.

\begin{table*}[!h]
\small
\centering
\begin{tabularx}{\linewidth}{Xcccccccccccccccc}
    \toprule
    \multirow{2}{*}{Model} & \multicolumn{3}{c}{Agent Name} & \multicolumn{3}{c}{Reference Number} & \multicolumn{3}{c}{Group Number}\\
    \cmidrule(l){2-4}
    \cmidrule(l){5-7}
    \cmidrule(l){8-10}
    \cmidrule(l){11-12}
    
      & Precision & Recall & F1 & Precision & Recall & F1 & Precision & Recall & F1 \\  \midrule
    
    XGBoost               & 0.9570  & 0.6617 & 0.7824 & 0.9636 & 0.8598 & 0.9088 & 0.9749 & 0.8969 & 0.9343 \\
    XGBoost + AED          & 0.9494  & 0.7634 & 0.8463 & 0.9732 & 0.8637 & 0.9152 & 0.9532 & 0.7523 & 0.8409 \\
    XGBoost + AEC          & 0.9567 & 0.6682 & 0.7868 & 0.9739 & 0.8506 & 0.9081 & 0.9562 & 0.6605 & 0.7813 \\
    XGBoost + AED + AEC & 0.9508  & 0.7569 & 0.8429 & 0.9689 & 0.8773 & 0.9208 & 0.9531 & 0.7405 & 0.8335 \\
    \hdashline
    Gemini 1.5             & 0.9563  & 0.8472 & 0.8985 & 0.9499 & 0.7541 & 0.8408 & 0.9796 & 0.6656 & 0.7927 \\
    Gemini 1.5 + AEC       & 0.9602  & 0.8011 & 0.8734 & 0.9569 & 0.7221 & 0.8231 & 0.9815 & 0.5979 & 0.7431 \\
    GPT 3.5                & 0.9373  & 0.8829 & 0.9093 & 0.9355 & 0.7953 & 0.8598 & 0.9493 & 0.9508 & 0.9500 \\
    GPT 3.5 + AEC          & 0.9415  & 0.8626 & 0.9003 & 0.9432 & 0.8138 & 0.8737 & 0.9506 & 0.9574 & 0.9540 \\
    \rowcolor{gray!20} Fine-tuned GPT 3.5 + AEC    & \multirow{2}{*}{0.9192}  & \multirow{2}{*}{0.9985} & \multirow{2}{*}{\textbf{0.9572*}} & \multirow{2}{*}{0.9386} & \multirow{2}{*}{0.9942} & \multirow{2}{*}{\textbf{0.9656*}} & \multirow{2}{*}{0.9556} & \multirow{2}{*}{0.9933} & \multirow{2}{*}{\textbf{0.9741*}} \\
    \bottomrule
\end{tabularx}
\captionsetup{font=footnotesize}
\caption{Model performance for the \textit{Direct Verification} setting in correctly reviewing Agent Name, Reference Number, and Group Number. Fine-tuned GPT 3.5 + AEC refers to the model fine-tuned for auto-reviewing using corrected transcripts. The results highlighted in gray are from the fine-tuned model, all other models have not been fine-tuned. (AED: ASR Error Detection, AEC: ASR Error Correction, GPT 3.5: GPT 3.5 Turbo). McNemar’s tests were conducted on the best-performing model for each field against its baseline (XGBoost), and all comparisons showed statistically significant improvements ($*: p<0.001$)}
\label{tab:direct_verification}
\end{table*}

\subsection{Error Patterns}
A major source of incorrect predictions at this stage stems from errors in the call transcripts, which can result in either incorrect field values being approved or correct ones being missed.

Our task faces two main challenges: 1) detecting errors in call-level field extraction, which is a highly imbalanced classification problem, and 2) auto-correcting detected errors, which requires understanding ASR error patterns. One common error pattern involves similar pronunciations, such as a mistranscribed reference number (Rina A 01012024 instead of Sabrina A 01012024). Another common issue arises from inaccurate long sequence transcripts, such as missing or redundant digits (e.g., `10001234' missing a 0, or `1234560' with an extra 0). These ASR errors present a bottleneck for the auto-review process.

\section{Error Handling}
Traditional ASR error correction models aim to detect and correct all errors in a transcript~\citep{lu2019impact}. In contrast, our focus is not on correcting grammatical errors, but on ensuring the accuracy of the information relevant to benefit verification. As noted in recent studies~\citep{zhu2021improving}, using n-best alternatives significantly improves error correction. 
In our experiments, providing multiple transcript alternatives improves data extraction performance.
Therefore, we use n-alternatives at both the pseudo-label generation and error correction stages\footnote{Please refer to \ref{sec:prelim_audio_gemini} and \ref{app:asr_alternatives} for more details about our main model architecture decision.}

\begin{figure}
    \centering
    \includegraphics[width=0.50\textwidth]{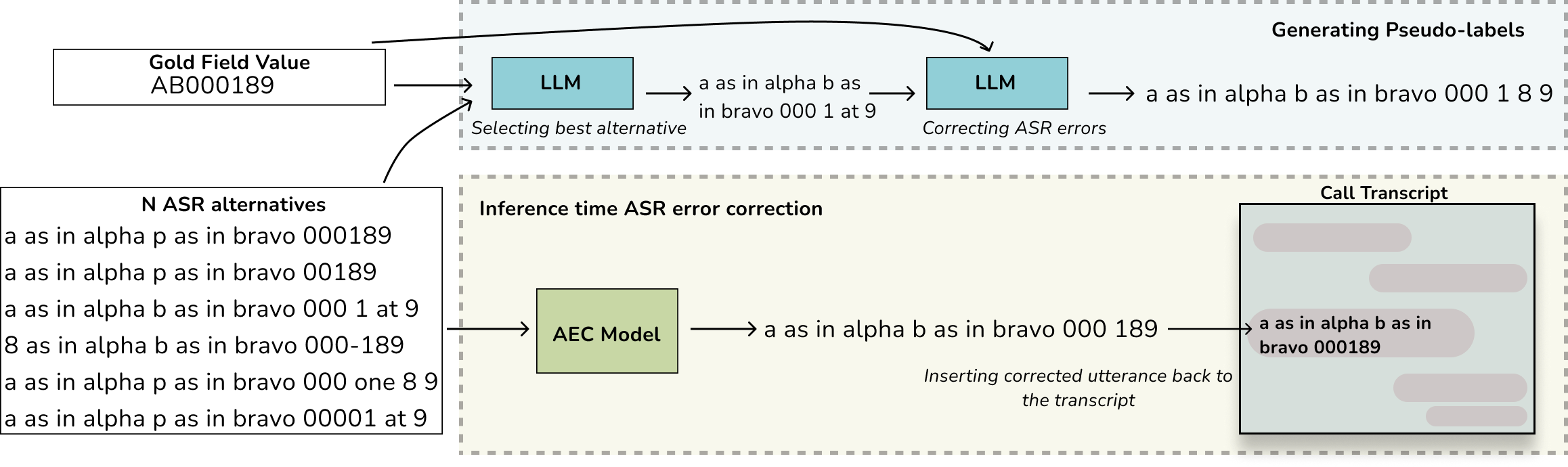}
    \captionsetup{font=footnotesize}
    \caption{An overview of the ASR error handling component. $n$ ASR alternatives are used to generate the pseudo-labels that are then used for training the AEC model. During inference, the corrected utterances are inserted back into the transcript.}
    \label{fig:overview}
\end{figure}

\subsection{Generating Pseudo-Labels}
\begin{algorithm}[tb]
\small
\captionsetup{font=small}
\caption{Correcting ASR transcript using gold field value}
\label{alg:asr_correction}
\begin{algorithmic}[1]
\STATE \textbf{Input}: $ASR_N$ (list of ASR alternatives), $field_{gold}$ (corrected field value)  \\

\STATE $ASR_{best}$ $\gets$ $f_{best\_alternative}^{LLM}$($ASR_N$, $field_{gold}$)
\STATE $ASR_{corr}$ $\gets$ $f_{correct\_transcript}^{LLM}$($ASR_{best}$, $field_{gold}$)

\STATE \textbf{return} $ASR_{corr}$
\end{algorithmic}
\end{algorithm}

Manually curating an error correction dataset from a large number of calls is expensive and time-consuming. Instead, we leverage existing ASR transcripts and human-reviewed field values from past calls to create a specialized dataset for error correction.

To generate pseudo-labels, we prompt an LLM to correct noisy transcripts so that the information aligns with the gold field value. In initial experiments, we found that when multiple errors were present in a transcript\footnote{The best transcript returned by the ASR model may not be the most accurate for benefit verification.}, the LLM struggled to correct all of them. To address this, we use n-alternatives and break pseudo-label generation into two steps. First, we provide the LLM\footnote{We use the Gemini model for generating pseudo-labels.} with all n-alternatives and the gold field value, asking it to choose the best alternative, we formalize this as $f_{best\_alternative}^{LLM}$($ASR_N$, $field_{gold}$). Then, using the selected alternative and the gold value, we prompt the LLM again to correct the transcript, we call this function $f_{correct\_transcript}^{LLM}$($ASR_{best}$, $field_{gold}$).~\autoref{fig:overview} gives the workflow on pseudo-label generation. In all experiments, we set $n = 10$~\footnote{Additional details on the choice of $n$ are given in appendix~\ref{app:asr_alternatives}}. Detailed prompts are described in~\autoref{sec:model_prompts}, and the algorithm for locating utterances is presented in~\autoref{sec:utts_extracted}.

\subsection{Automatic Error Correction Model}

For the ASR Error Correction (AEC) model, we use Mistral~\citep{jiang2023mistral} as the base model for error handling tasks\footnote{We chose Mistral due to its open-source availability and, in our preliminary experiments with random subset samples, performed better than LLaMA-8B-instruct.}.
The AEC model focuses exclusively on correcting utterances containing key field values.
We first isolate those utterances for each field type. The corresponding pseudo-labels are generated only during the training phase. We provide \textit{n} alternatives as input to the model and train using the pseudo-labels. Given the \textit{n} alternatives, the AEC model is trained to output a single correct transcript. After the correction, the corrected utterances are inserted back to their original place in the full call transcript.

\subsection{Automatic Error Detection Model}
Error detection can be considered a component of the full auto-correction pipeline~\citep{fang-etal-2022-non, leng2023softcorrect} and can be easily integrated into various ML models as an additional feature. To assess its impact, we examine the effect of incorporating a simple error detection signal into our production-level model.

The ASR Error Detection (AED) model is trained similarly to the AEC model but differs in its output. Instead of generating a corrected transcript, the AED model produces a binary classification: \textit{True} if the first of the n alternatives is noisy and \textit{False} otherwise. To adapt the AEC training data for this task, we label an instance as \textit{True} if the best alternative differs from the pseudo-corrected transcript and \textit{False} otherwise.


\section{Results}

\subsection{Evaluation Setting}
The goal of both \textit{Direct Extraction} and \textit{Direct Verification} is to determine whether a given live-call field value is correct. If the gold field value is the same as the live-call value and the model predicts it as correct, we consider that a correct prediction.
Since our primary focus is on `auto-approval', we evaluate results specifically for that class.

Given the dataset's high imbalance, we report precision, recall, and F1 scores. For \textit{Direct Extraction}, we also measure exact match and normalized edit distance. The baseline in both evaluation settings is the model that is just provided the best ASR transcript, without any error correction~\footnote{We measure the efficacy of the error correction model by evaluating directly on the downstream task of benefit verification as opposed to intrinsic evaluation metrics such as ROUGE, since we do not have gold corrected transcripts.}.

\begin{table*}[!htb]
\centering
\small
\begin{tabular}{lccccc}
    \toprule 
    \textbf{Field Value} & \textbf{Precision$\uparrow$} & \textbf{Recall$\uparrow$} & \textbf{F1$\uparrow$} & \textbf{Accuracy$\uparrow$} & \textbf{NED$\downarrow$} \\ 
    \midrule
    \multicolumn{6}{c}{\textbf{Gemini}} \\ \midrule
    Agent Name & 0.9756 & 0.4568 & 0.6223 & 0.4403 & 0.2263 \\
    Reference Number & 0.9791 & 0.2958 & 0.4544 & 0.2785 & 0.4083 \\
    Group Number & 0.9942 & 0.3508 & 0.5186 & 0.3475 & 0.2673 \\
    Average & 0.9830 & 0.3746 & 0.5318 & 0.3554 & 0.3006 \\
    \midrule
    \multicolumn{6}{c}{\textbf{Gemini + AEC}} \\ \midrule
    Agent Name & 0.9776 & 0.4772 & 0.6413 & 0.4594 & 0.2187 \\
    Reference Number & 0.9787 & 0.3574 & 0.5236 & 0.3383 & 0.3822 \\
    Group Number & 0.9916 & 0.4262 & 0.5961 & 0.4248 & 0.2292 \\
    \textbf{Average} & \textbf{0.9823} & \textbf{0.4203} & \textbf{0.5870} & \textbf{0.4075} & \textbf{0.2767} \\
    \bottomrule
\end{tabular}
\captionsetup{font=footnotesize}
\caption{Performance metrics in the \textit{Direct Extraction} setting. `Gemini' is the baseline that only gets the best ASR transcript while `Gemini+AEC' gets the corrected transcript as input. \textit{NED: Normalized Edit Distance}}
\label{tab:direct_extraction}
\end{table*}

\subsection{Base Models}
For off-the-shelf LLMs, we report results on GPT~\citep{brown2020language,achiam2023gpt} and Gemini~\citep{team2023gemini} models with noisy ASR transcripts as baseline and after performing error correction. The detailed prompts can be found in~\autoref{sec:model_prompts}.
We also integrate the AEC model into the auto-review model used in a feature-based model architecture. We use XGBoost model architecture so we can leverage all of the statistical and historical features\footnote{Features include textual features extracted from live-call field values (e.g., regular expression patterns for expected formats for each field), call STT transcripts and statistical and historical features extracted from benefit verification client and call recipient insurance company.} and LLM models (e.g., field value extractions using LLMs) as features for making final auto-approval decisions. We do not compare against other specialized error correction models, as they either focus on grammatical error correction~\citep{li2024detection, ma2023n} or rely on specialized knowledge graphs~\citep{discriminative-entity-aware-language} or manual annotations.

\subsection{Analysis}

Our goal is to assess the impact of ASR error correction on the overall performance of the \textbf{Auto Review} pipeline.
Ultimately, the choice of model depends on the specific use case and the acceptable trade-off between precision and recall.

\paragraph{Direct Verification}~\autoref{tab:direct_verification} presents the results for direct verification. We first examine the XGBoost model within the feature-based pipeline. Adding a simple binary feature for AED (indicating whether the transcript is noisy) improves performance for two out of three fields. Further incorporating corrected transcripts, the `XGBoost + AED + AEC' model significantly enhances the F1 score for `Agent Name' (0.7824$\rightarrow$0.8428) and achieves the best performance on `Reference Number' (0.9088$\rightarrow$0.9208).
The `Gemini 1.5 + AEC' model improves precision across all fields but at the cost of reduced recall. In contrast, `GPT 3.5 + AEC' enhances overall performance across all fields, except for a slight recall drop in `Agent Name'. Notably, it achieves the highest accuracy for `Group Number'. Fine-tuned GPT model with AEC obtained the highest F1 score on all fields by improving the recall substantially but resulted in a lower precision.
Compared to LLMs, XGBoost models achieve higher precision but lower recall. This is due to their reliance on specialized regular expressions for field formats~\footnote{e.g., predefined patterns for group numbers, reference numbers, and agent name capitalization} as well as historical and statistical features. However, these constraints limit their generalization to diverse cases.

\paragraph{Direct Extraction} 
Unlike the direct verification approach, the AEC model does not receive the live-call extracted field value as input. Instead, it extracts the field value directly from the ASR transcript. This extracted value is then compared to the live-call field values as an additional validation step. If both values match, the system auto-approves the result; otherwise, it requests a second human review.
As shown in~\autoref{tab:direct_extraction}, this method results in lower recall, as the model often fails to approve correct values due to variations in ASR outputs. For instance, as illustrated in ~\autoref{fig:example_snapshot}, the \textit{direct verification} model may approve the live-call group number despite minor errors in the transcript (e.g., ignoring an incorrect `8'). In contrast, the \textit{direct extraction} model may output alternative values such as `8D0156' or `AD0156', increasing susceptibility to ASR errors.
However, this approach achieves significantly higher precision. After applying ASR error correction, precision remains stable across all fields, while recall improves substantially, yielding an average F1 score improvement of \textbf{5.5\%}. While failing to auto-approve correct values is undesirable, it is preferable to approving incorrect extractions and passing them to customers.

A hybrid model combining both settings could be implemented in production. \textit{Direct verification} would be applied to less critical fields~\footnote{Critical fields are those where incorrect values can have a significant negative impact on customers.}, leading to a higher overall F1 score and saving time on review. \textit{Direct extraction} would be reserved for critical fields, approving them under a more stringent setting.

\section{Conclusion}
We introduced \textbf{Auto Review}, a two-stage pipeline that enhances information extraction from healthcare phone calls. Our approach reduces human verification while maintaining high accuracy. 
The second stage involves an ASR error correction framework, leveraging n-best ASR alternatives to generate pseudo-labels for training an error correction model.
This framework is adaptable across domains, provided some past manually reviewed data is available.
Results show that ASR error correction improves precision and recall across key fields, with \textit{Direct Verification} offering higher recall and \textit{Direct Extraction} achieving higher precision.

The results reported in this paper reflect the isolated performance of a model component within a larger production system. In real-world deployment, additional pipeline components—including human-in-the-loop mechanisms and cross-field verification models—contribute to significantly higher precision. This underscores the complementary role of system-level engineering in achieving production-grade performance alongside core model development.

\section{Ethical Statement}
All experiments described in this paper were conducted in compliance with applicable privacy and data protection regulations. Specifically, interactions with third-party models, including OpenAI’s GPT-3.5 Turbo and Google’s Gemini, were governed by appropriate Business Associate Agreements (BAAs) if required under the Health Insurance Portability and Accountability Act (HIPAA). These controls were designed to ensure that no Protected Health Information (PHI) was exposed to external service providers for training or other purposes beyond our immediate use case, and that at no point was PHI stored in third-party companies or used to improve or fine-tune the third-party models themselves.

For model inferences in our main experiments with GPT-3.5 Turbo and Gemini 1.5 Pro APIs, the total estimated cost was \$303, based on publicly available pricing at the time of experimentation. This included approximately \$260 for Gemini 1.5 Pro with audio input, \$20 for Gemini 1.5 Pro with text input, and \$23 for GPT-3.5 Turbo (16k context) with text input.



\appendix

\section{Training Description}\label{sec:training_stats}
For the AEC model, we use Mistral-7B-Instruct-v0.3, which was trained with a batch size of 16, gradient accumulation step set to 2 using 1 A100 GPU. Training took ~9 hours.
In all our AEC experiments, the number of alternatives, \textit{n}, is fixed to 10. LoRA~\citep{hu2022lora} is used for parameter-efficient training using the LLaMA-Factory library~\citep{zheng2024llamafactory}.
We use the Gemini 1.5 model to generate the pseudo-labels. Google STT model is used as the base STT model for all ASR transcripts\footnote{In preliminary experiments, we found fine-tuning ASR helped improving the general performance metric such as word error rate (WER) but observed the similar issues especially from unseen field values. See more details in \ref{sec:asr_preliminary}}. 

\section{Relevant Utterance Isolation}\label{sec:utts_extracted}
Algorithm ~\autoref{alg:field_trigger} presents the algorithm to isolate only those utterances from the call transcripts that are highly likely to contain the field value information we want to extract. It starts collecting agent utterances after the conversational AI model asks for information regarding that field, those trigger questions are pre-defined and passed to the algorithm in field\_triggers.

\begin{algorithm}
\caption{Extract Utterances for Fields of Interest}
\label{alg:field_trigger}

\begin{algorithmic}[1]
\REQUIRE call\_transcript (list of tuples with speaker and utterance), field\_triggers (list of trigger utterances)

\STATE Initialize an empty list $agent\_responses$
\STATE Set $collect\_responses \gets \textbf{false}$

\FOR{each $(speaker, utterance)$ in $call\_transcript$}
    \IF{not $collect\_responses$ \AND $utterance$ contains any phrase in $field\_triggers$}
        \STATE $collect\_responses \gets \textbf{true}$
    \ELSIF{$collect\_responses$}
        \IF{$speaker = \text{Agent}$}
            \STATE Append $utterance$ to $agent\_responses$
        \ELSIF{$speaker = \text{AI Model}$}
            \STATE $collect\_responses \gets \textbf{false}$
        \ENDIF
    \ENDIF
\ENDFOR

\RETURN $agent\_responses$

\end{algorithmic}
\end{algorithm}

\section{Preliminary Experiments}
\subsection{Experiments with Gemini using Audio Input}\label{sec:prelim_audio_gemini}

\begin{table}[!htb]
\centering
\small
\begin{tabular}{lccc}
    \toprule
    \textbf{Field Value} & \textbf{Precision} & \textbf{Recall} & \textbf{F1} \\
    \midrule
    \multicolumn{4}{c}{\textbf{Gemini with Audio}} \\ \midrule
    Agent Name & 0.9838 & 0.1205 & 0.2148 \\
    Reference Number & 0.9816 & 0.3875 & 0.5556 \\
    Group Number & 0.9965 & 0.4323 & 0.6030 \\
    \midrule
    \multicolumn{4}{c}{\textbf{XGBoost Model}} \\ \midrule
    Agent Name & 0.9570 & 0.6617 & 0.7824 \\
    Reference Number & 0.9636 & 0.8598 & 0.9088 \\
    Group Number & 0.9749 & 0.8969 & 0.9343 \\
    \bottomrule
\end{tabular}
\captionsetup{font=footnotesize}
\caption{Performance metrics for Agent Name, Reference Number, and Group Number in the \textit{Direct Extraction} setting using Gemini with audio input. The audio-based model suffers from very low recall.}
\label{tab:direct_extraction}
\end{table}

\begin{figure*}[t]
    \centering
    \captionsetup{font=footnotesize}
    \includegraphics[width=\textwidth]{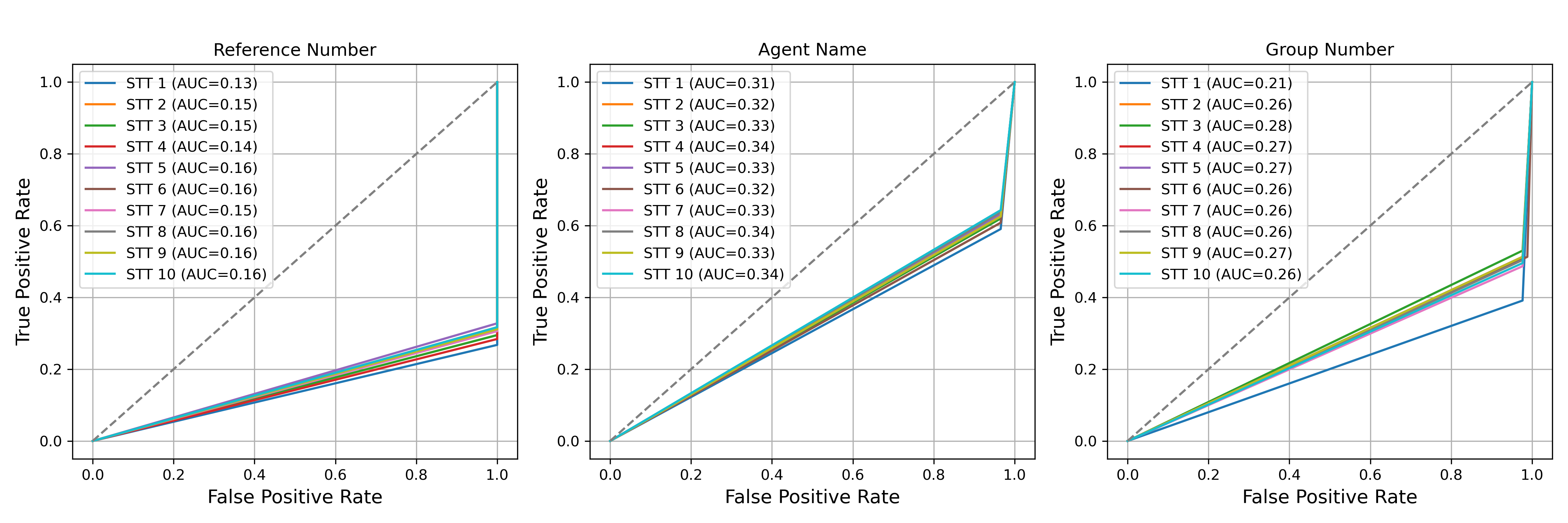}
    \captionsetup{font=footnotesize}
    \caption{The ROC curves for the three field types when using different numbers of transcript alternatives as input. The Gemini model is provided the transcript to extract the field value, which is then compared with the gold field value. Providing multiple alternatives improves performance.}
    \label{fig:roc}
\end{figure*}

For our preliminary analysis, we experimented with off-the-shelf multimodal LLM (Gemini 1.5) with the same prompt we used for ASR text transcript direct extraction (\autoref{tab:prompts_extraction_GN}, \autoref{tab:prompts_extraction_AN}) except for the instruction which tells to use the attached audio instead of the text providing the call audio recording. Gemini obtained high precision overall, but its recall is too low to effectively reduce human review time in industry settings with a large number of concurrent phone calls. When we analyzed false positive auto-approved samples, it made similar mistakes with ASR models incorrectly adding 0 or missing a few digits for long alphanumeric field values or misspelling rare agent names with more common names. Thus, we designed ASR error detection and correction models focusing on the field values of the data types that are highly vulnerable to such errors and cannot be resolved by off-the-shelf LLMs or other feature-based models.

\subsection{Experiments with ASR systems}\label{sec:asr_preliminary}
We conducted preliminary experiments using Google STT and Whisper (Whisper Large V3\footnote{\url{https://huggingface.co/openai/whisper-large-v3}}) to choose the most suitable ASR system for our field value output extraction tasks. Although Google STT obtained a higher performance than Whisper, it was not available for fine-tuning so we fine-tuned Whisper model using the subset of our full data to explore the best ASR system options (785 outputs for training set, 390 outputs for validation set and 510 outputs for test set). We found that our fine-tuned Whisper model improved the general evaluation metrics but we still observed similar issues with mistranscripts with digits or letters missing for long sequence outputs; especially with the patterns which did not exist in training set (see more details in Table~\ref{tab:asr_comparison}). Thus, collecting ground truth labels for all such cases required a large human labeling effort and it was not scalable for our task with real world data so we chose off-the-shelf Google STT for our main experiments.

\begin{table}[!htb]
\centering

\begin{tabular}{l|l|ll}
    \hline
    ASR System & Word Error Rates & Norm. Edit \\
    \hline
    Google STT & 0.602 & 0.430    \\
    Whisper & 0.757 & 0.485  \\
    FT Whisper & 0.349 & 0.216 \\
    \hline
\end{tabular}
\captionsetup{font=footnotesize}
\caption{Performance metrics for ASR systems on task output transcription. FT Whisper: fine-tuned Whisper, Norm. Edit: normalized edit distance (edit distance between the transcript and the ground truth divided by the maximum value among the lengths of the two).}
\label{tab:asr_comparison}
\end{table}

\begin{figure}[h]
    \centering
    \captionsetup{font=footnotesize}
    \includegraphics[width=0.5\textwidth]{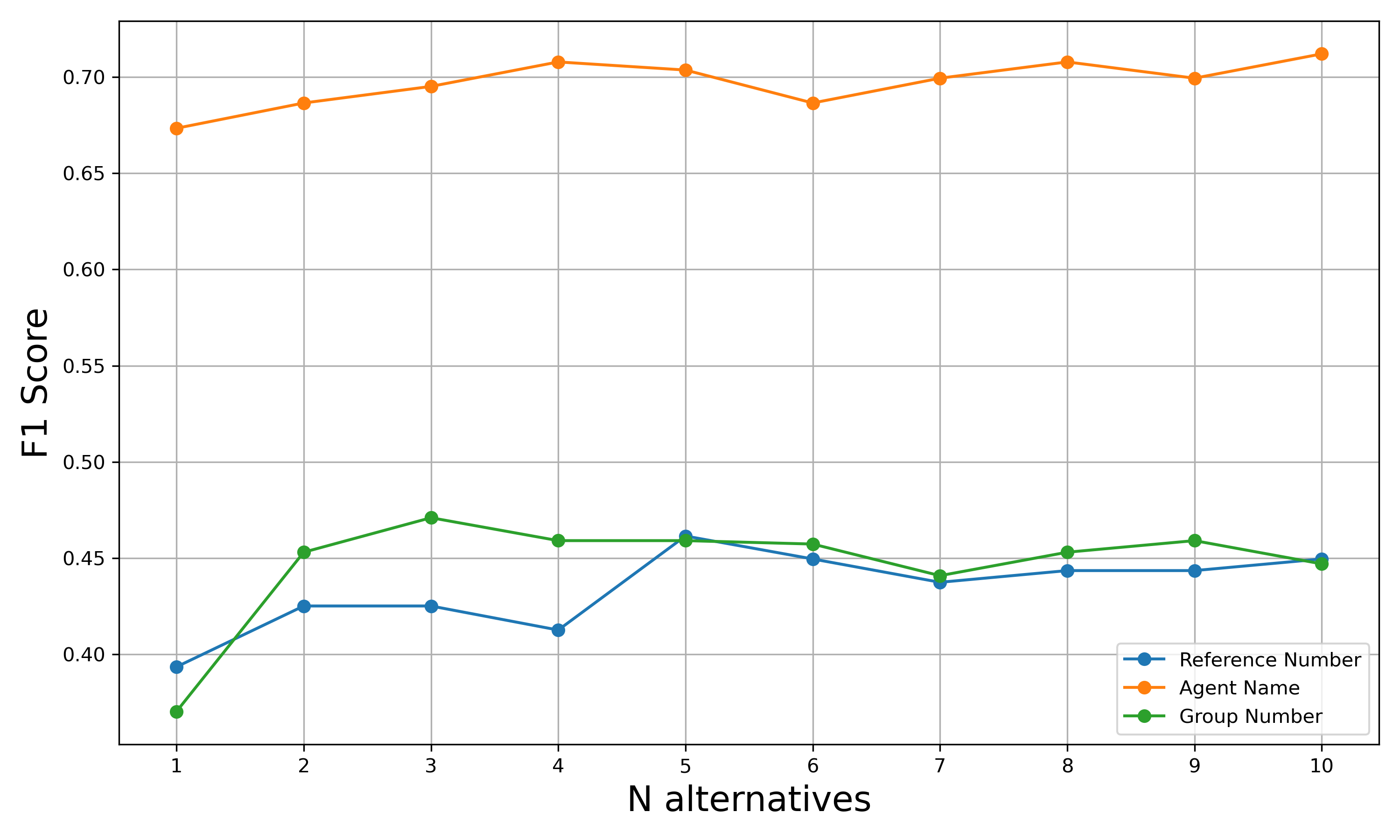}
    \captionsetup{font=footnotesize}
    \caption{F1 scores for LLM performance across the three field types, based on correctly extracting field values from transcripts. The input transcript to the LLM includes multiple ASR alternatives. A significant performance improvement is observed when incorporating multiple alternatives instead of relying solely on the best one.}
    \label{fig:f1}
\end{figure}

\subsection{Number of ASR Alternatives}
\label{app:asr_alternatives}
We conducted experiments to assess the impact of using multiple ASR alternatives on field value extraction. Using a subset of 200 calls, we measured LLM performance in extracting three key fields. Specifically, we prompted Gemini to extract field values from transcripts while varying the number of ASR alternatives, with $n=1$ corresponding to using only the best transcript. The prompts for these experiments follow the \textit{Direct Extraction} approach and are detailed in~\autoref{tab:prompts_extraction_AN} and~\autoref{tab:prompts_extraction_GN}.
As shown in~\autoref{fig:f1} and~\autoref{fig:roc}, incorporating multiple ASR alternatives significantly improves performance across all field values.
Since the optimal value of $n$ varies by field type and we want to train a single cohesive AEC model, we chose $n=10$ in all our experiments.

\section{Model Prompts}\label{sec:model_prompts}
The prompts used for all the experiments are given below. The in-context examples used in the experiments have been removed because they contain sensitive patient information.

\begin{table*}[!htb]
\footnotesize
\centering
\begin{tabularx}{\linewidth}{C} 
    \toprule
    \textbf{<INSTRUCTIONS>} You are ``\{our conversational AI model name\}'', a digital assistant calling a healthcare insurance company to get benefits information for a member. Given the STT transcript of phone conversations between you and the health insurance company agent, check if all of your answers to the given questions are correct. Please respond using ``correct'' or ``incorrect'', checking whether all the answers to the questions in the call are correct or not, and provide your reasoning in JSON format. Here are example cases for each answer: \newline
    1. ``correct'': select this option only if all the answers are correct based on the call transcript. \newline
    2. ``incorrect'': select this option if you see any of the answers to the questions is incorrect. \newline
    Below are sample responses and reasons: \newline
    Reason: Among 4 questions asked, the answer to the second question should have been ``True''. // Your response: \{``response'': ``incorrect''\} \newline
    Reason: All of the answers to the given 5 questions are correct. // Your response: \{``response'': ``correct''\} \newline
    Reason: There was one question and the agent could not provide the answer and the answer was ``agent did not provide this information''. // Your response: \{``response'': ``correct''\} \textbf{</INSTRUCTIONS>} \newline
    
    \textbf{<TARGET\_QUESTION\_GUIDELINES>} Some additional guidelines for specific questions with examples for the questions of ``agentName'', ``referenceNumber'', and ``groupNumber'': \newline
    1. Note if the agent spells it out or uses nato alphabet. For example, if the agent says ``c as in Charlie 2 n as in Nancy 3 c as in Tango G is in gold'', you should collect ``C2N3TG''. With STT mistranscriptions, you should follow the nato alphabet over the spelling. \newline
    2. Unless there is a word or name used, capitalize all letters and remove any spaces. For example, if the agent says ``group number is 123 456 789'', you should collect ``1234567890''. \newline
    3. There might be speech to text transcription errors (e.g. ``8'' instead of "H" or ``for'' instead of ``4'') For example, they might say ``C like Tango'' and in this case you should get the spelling to include T, not C.
    \textbf{</TARGET\_QUESTION\_GUIDELINES>} \newline
    
    \textbf{<TARGET\_QUESTION\_EXAMPLES>}
    [reason // questions // your response] \newline
    - Reason: ``the agent spelled out their name as Jane and said C like Tango''
    Question: ``Question 1: agentName? Answer: 'Jane T''' // Your response: \{\{``response'': ``correct''\}\} Reason: ``the agent gave their name as Jane and said his last name initial is O as in Oscar and said there were no reference numbers'' // Question:  "Question 1: agentName? Answer: 'Jane O'. Question 2: referenceNumber? Answer: 'Jane O 06242024'" // Your response: \{\{``response'': ``correct''\}\} \newline 
    - Reason: ``the agent said t i a b for boy so likely the last name initial is B so the first name is Tia'' // Question ``agentName'': ``Tia B'', ``referenceNumber'': ``12345''\}\} \newline 
    - Reason: ``the agent said d a r a for alpha my initial so likely A is their last name initial so the first name is Dar'' // Question: "Question 1: agentName? 'Dar A''' // Your response: \{\{``response'': ``correct''\}\} \newline
    - Reason: "the agent said their name was j a qu a i d i a last initial K so their name is Jaquaidia K and they said the reference number was their name and the date" // Question: "Question 1: agentName? 'Jaquaidia K'. Question 2: referenceNumber? 'Jaquaidia K 06012024'// Your response: {{"response": "correct"}} \newline 
    - Reason: "the agent said their name was Jasmine but spelled it out as J A S M I N so with that spelling their name must be Jasmin" // Question: "Question 1: agentName? 'Jasmine'" // Your response: \{\{``response'': ``incorrect''\}\}
    - Reason: ``the agent said their name was Sam but spelled it out as s a m y r so with that spelling their name must be Samyr'' // Question: ``Question 1: agentName? 'Samyr''' // Your response: \{\{``response'': ``correct"\}\} \newline 
    - Reason: ``the agent spelled their name as 'p as in paul n as in nancy o t t r i c last initial is d' so their name is Pnottric D and gave no reference number" // Question: ``Question 1: agentName? 'Pnottric D'. Question 2: referenceNumber? 'Pnottric D 06012024''' // Your response: \{\{``response'': ``correct''\}\}
    
    \textbf{</TARGET\_QUESTION\_EXAMPLES>}

    Below is the STT transcript of the call.
    
    \textbf{[transcript]} \newline

    Answer if all of the following questions and answer pairs are correct in the JSON format as in the example in the instruction
    \newline
    \textbf{[question\_answer\_pairs]}
    \\
    \bottomrule
\end{tabularx}
\caption{\textit{Direct Verification} prompt used for all fields.}
\label{tab:prompts_verification}
\end{table*}

\begin{table*}[!htb]
\footnotesize
\centering
\begin{tabularx}{\linewidth}{C} 
    \toprule    
    \textbf{</INSTRUCTIONS>} You are a capable annotator who can identify and correct issues in STT transcript.
    You will be given alternative STT transcripts and corresponding extracted name. 
    Pick the best alternative that most correctly corresponds to the given extracted name.
    The best alternative is defined as: The alternative transcript from which we should be able to extract the name that matches the given extracted name.
    If there are multiple names present, usually we only care about the last name. Ignore the name ``\{our conversational AI model name\}'' if it is present in the transcript.
    The alternative transcripts are separated by ``\#''.
    Give the output in json format of \{\{``Output'': best_transcript\}\}
    
    \textbf{</INSTRUCTION>} \newline
    \textbf{<EXAMPLES>} \newline
    Here are some examples of the STT transcripts along with the extracted value and the outputs separated by ``//'' (i.e., STT transcripts, extracted name // your output):\newline
    [Examples]\newline
    \textbf{</EXAMPLES>}
    
    Now provide your answer from the following STT transcripts and extracted value:\newline
    [Input]
    \\
    \bottomrule
\end{tabularx}
\caption{Pseudo-label generation prompt for selecting the best alternative.}
\label{tab:prompts_PL_alternative}
\end{table*}

\begin{table*}[!htb]
\footnotesize
\centering
\begin{tabularx}{\linewidth}{C} 
    \toprule
    \textbf{<INSTRUCTIONS>} You are a capable annotator who can identify and correct issues in STT transcript.
    You will be given STT transcript and corresponding extracted value. 
    If the transcript is correct, you will simply return the transcript and if the transcript is wrong compared to the correctly extracted value, you need to correct the transcript appropriately.
    Pay special attention to the number of zeros in the extracted value and compare with the noisy transcript. Do not capitalize letters in the transcript if they are not originally capitalized, even if the extracted value has capitalized letters.
    Give the output in json format of \{\{``Output'': corrected_transcript\}\}
    
    \textbf{</INSTRUCTIONS>} \newline
    \textbf{<EXAMPLES>}
    Here are some examples of the STT transcript along with the extracted value and the outputs separated by ``//'' (i.e., STT transcript, extracted value // your output): \newline
    [Examples]\newline
    \textbf{</EXAMPLES>}
    
    Now provide your answer from the following STT transcript and extracted value:
    [Input]
    \\
    \bottomrule
\end{tabularx}
\caption{Pseudo-label generation prompt for error correction.}
\label{tab:prompts_PL_correction}
\end{table*}

\begin{table*}[!htb]
\footnotesize
\centering
\begin{tabularx}{\linewidth}{C} 
    \toprule
    \textbf{<PROMPT>}You are a capable annotator who can identify and correct issues in ASR transcript.
    You will be given a list of noisy ASR outputs, separated by ``\#''. Output the best possible ASR alternative.
    In some cases, the correct output will be one of the provided alternatives, in other cases you will have to identify patterns across the alternatives and output a cohesive correct transcript. \newline
    \textbf{</PROMPT>} \newline
    [Input]
    \\
    \bottomrule
\end{tabularx}
\caption{Automatic error correction model prompt.}
\label{tab:prompt_AEC}
\end{table*}

\begin{table*}[!htb]
\footnotesize
\centering
\begin{tabularx}{\linewidth}{C} 
    \toprule
    \textbf{<INSTRUCTIONS>}Given a transcript, extract the underlying group number value.
    Give the output in json format of \{\{``Output'': extracted value\}\}

    \textbf{</INSTRUCTIONS>} \newline
    \textbf{<EXAMPLES>}
    Here are some examples of the transcript along with the extracted output separated by ``//'' (i.e., text // your output):\newline
    [Examples]\newline
    \textbf{</EXAMPLES>}\newline
    Now provide your answer from the following text: \newline
    [Input]
    \\
    \bottomrule
\end{tabularx}
\caption{Direct extraction prompt for Group Number.}
\label{tab:prompts_extraction_GN}
\end{table*}

\begin{table*}[!htb]
\footnotesize
\centering
\begin{tabularx}{\linewidth}{C} 
    \toprule
    \textbf{<INSTRUCTIONS>} Given a transcript, extract the underlying name. Ignore ``\{our conversational AI model name\}'' if it appears in the transcript.
    If there are multiple names, extract the last one. Capitalize the first name initial and last name initial.
    Give the output in json format of \{\{``Output'': extracted value\}\}

    \textbf{</INSTRUCTIONS>} \newline
    \textbf{<EXAMPLES>}
    Here are some examples of the transcript along with the extracted output separated by ``//'' (i.e., text // your output):\newline
    [Examples]\newline
    \textbf{</EXAMPLES>}\newline
    Now provide your answer from the following text:\newline
    [Input]
    \\
    \bottomrule
\end{tabularx}
\caption{\textit{Direct Extraction} prompt for Agent Name and Reference Number.}
\label{tab:prompts_extraction_AN}
\end{table*}

\begin{table*}[!htb]
\footnotesize
\centering
\begin{tabularx}{\linewidth}{C} 
    \toprule
    \textbf{<INSTRUCTIONS>} You are ``\{our conversational AI model name\}'', a digital assistant calling a healthcare insurance company to get benefits information for a member. Given the STT transcript of phone conversations between you and the health insurance company agent, check if all of your answers to the given questions are correct. Please respond using ``correct'' or ``incorrect'', checking whether all the answers to the questions in the call are correct or not, and provide your reasoning in JSON format. Here are example cases for each answer: \newline
    1. ``correct'': select this option only if all the answers are correct based on the call transcript. \newline
    2. ``incorrect'': select this option if you see any of the answers to the questions is incorrect. \newline
    Below are sample responses and reasons: \newline
    Reason: Among 4 questions asked, the answer to the second question should have been ``True''. // Your response: \{``response'': ``incorrect''\} \newline
    Reason: All of the answers to the given 5 questions are correct. // Your response: \{``response'': ``correct''\} \newline
    Reason: There was one question and the agent could not provide the answer and the answer was ``agent did not provide this information''. // Your response: \{``response'': ``correct''\} \textbf{</INSTRUCTIONS>} \newline
    
    \textbf{<TARGET\_QUESTION\_GUIDELINES>} Some additional guidelines for specific questions with examples for the questions of ``agentName'', ``referenceNumber'', and ``groupNumber'': \newline
    1. Note if the agent spells it out or uses nato alphabet. For example, if the agent says ``c as in Charlie 2 n as in Nancy 3 c as in Tango G is in gold'', you should collect ``C2N3TG''. With STT mistranscriptions, you should follow the nato alphabet over the spelling. \newline
    2. Unless there is a word or name used, capitalize all letters and remove any spaces. For example, if the agent says ``group number is 123 456 789'', you should collect ``1234567890''. \newline
    3. There might be speech to text transcription errors (e.g. ``8'' instead of "H" or ``for'' instead of ``4'') For example, they might say ``C like Tango'' and in this case you should get the spelling to include T, not C.
    \textbf{</TARGET\_QUESTION\_GUIDELINES>} \newline
    
    \textbf{<TARGET\_QUESTION\_EXAMPLES>}
    [reason // questions // your response] \newline
    - Reason: ``the agent spelled out their name as Jane and said C like Tango''
    Question: ``Question 1: agentName? Answer: 'Jane T''' // Your response: \{\{``response'': ``correct''\}\} Reason: ``the agent gave their name as Jane and said his last name initial is O as in Oscar and said there were no reference numbers'' // Question:  "Question 1: agentName? Answer: 'Jane O'. Question 2: referenceNumber? Answer: 'Jane O 05012024'" // Your response: \{\{``response'': ``correct''\}\} \newline 
    - Reason: ``the agent said t i a b for boy so likely the last name initial is B so the first name is Tia'' // Question ``agentName'': ``Tia B'', ``referenceNumber'': ``12345''\}\} \newline 
    - Reason: ``the agent said d a r a for alpha my initial so likely A is their last name initial so the first name is Dar'' // Question: "Question 1: agentName? 'Dar A''' // Your response: \{\{``response'': ``correct''\}\} \newline
    - Reason: "the agent said their name was j a qu a i d i a last initial J so their name is Jaquaidia K and they said the reference number was their name and the date" // Question: "Question 1: agentName? 'Jaquaidia K'. Question 2: referenceNumber? 'Jaquaidia K 06012024'// Your response: {{"response": "correct"}} \newline 
    - Reason: "the agent said their name was Jasmine but spelled it out as J A S M I N so with that spelling their name must be Jasmin" // Question: "Question 1: agentName? 'Jasmine'" // Your response: \{\{``response'': ``incorrect''\}\}
    - Reason: ``the agent said their name was Sam but spelled it out as s a m y r so with that spelling their name must be Samyr'' // Question: ``Question 1: agentName? 'Samyr''' // Your response: \{\{``response'': ``correct"\}\} \newline 
    - Reason: ``the agent spelled their name as 'p as in paul n as in nancy o t t r i c last initial is g' so their name is Pnottric G and gave no reference number" // Question: ``Question 1: agentName? 'Pnottric G'. Question 2: referenceNumber? 'Pnottric G 06012024''' // Your response: \{\{``response'': ``correct''\}\}
    
    \textbf{</TARGET\_QUESTION\_EXAMPLES>}

    Below is the STT transcript of the call.
    
    \textbf{[transcript]} \newline

    Answer if all of the following questions and answer pairs are correct in the JSON format as in the example in the instruction
    \newline
    \textbf{[question\_answer\_pairs]}
    \\
    \bottomrule
\end{tabularx}
\caption{\textit{Direct Verification} prompt used for all fields.}
\label{tab:prompts_verification}
\end{table*}

\end{document}